# Gait Analysis for a Tiltrotor: The Dynamic Invertible Gait

Zhe Shen * and Takeshi Tsuchiya

Department of Aeronautics and Astronautics, The University of Tokyo, Tokyo 1138654, Japan;
tsuchiya@mail.ecc.u-tokyo.ac.jp
* Correspondence: zheshen@g.ecc.u-tokyo.ac.jp

**Abstract:** Conventional Feedback-Linearization-based controller, applied to the tilt-rotor (eight inputs), results in the extensive changes in the tilting angles, which are not expected in practice. To solve this problem, we introduce the novel concept "UAV gait" to restrict the tilting angles. The gait plan was initially to solve the control problems for quadruped (four-legged) robots. Transplanting this approach, accompanied by feedback linearization, to the tiltrotor may cause the well-known non-invertible problem in the decoupling matrix. In this research, we explore the invertible gait for the tiltrotor and apply feedback linearization to stabilize the attitude and the altitude. The equivalent conditions to achieve a full-rank decoupling matrix are deduced and simplified to a near zero roll and zero pitch. This paper proposed several invertible gaits to conduct the attitude–altitude control test. The accepted gaits within the region of interest are visualized. The experiment is simulated in Simulink, MATLAB. The results show the promising response in attitude and altitude.

## 1. Introduction

In the past decade, tiltrotors have gained great interest. The tiltrotor is a novel type of quadrotor [1–8] whose axes of the propellers are tilting, providing the ability to change the direction of each thrust. Typical control methods to stabilize a tiltrotor include LQR and PID [9–11], backstepping and sliding mode [12–16], feedback linearization [17–23], optimal control [24–26], adaptive control [27,28], etc. Among these, feedback linearization [29–31] explicitly decouples the nonlinear parts and enables utilizing the over-actuated properties. This approach is not only effective for the tiltrotor but also for the tiltrotor with predetermined tilting angles [32].

With such benefits of feedback linearization, the result provides behaviors such as tracking with a rapid response for sophisticated references [17,18,22]. However, several potential risks may hinder the application of this technique. One of them is the saturation restriction. This limit includes the upper bound of a given motor and the non-negative bound of the thrust; reaching a negative thrust is usually not acceptable in applications. Some research has aimed to avoid touching either of the above bounds [33–36]. Although hitting the bound does not necessarily mean that the result will be unstable, the corresponding stability criteria for these cases can be hard to trace or to generalize [37–39].

Another typical issue in feedback linearization applied in over-actuated systems is the so-called "State Drift" phenomenon in [40]; the state or input may drift as the simulation time extends. In [1,22], they define an optimal control problem to avert this problem. The extra requirements restrict the freedom of the inputs to some extent. Unfortunately, the relevant stability proof is not provided. An alternative way to avoid state drift is to reduce the number of inputs by predefining some of them [40]. This method is inspired by the gait plan, which is widely adopted in quadruped (four-legged) robots [41–45].

The number of inputs (the magnitude of each thrust and the direction of each thrust) of the tiltrotor in this research is eight at most. The number of degrees of freedom is six (attitude and position). The current feedback-linearization-based control strategies to stabilize this tiltrotor are categorized into two groups.

One is to manipulate all eight inputs so that the tiltrotor becomes an over-actuated system with the potential to maneuver all degrees of freedom simultaneously. The other is to sacrifice the number of inputs to make it equal to the number of degrees of freedom; a typical way is to make the tilting angles of the thrusts at the opposite arms equal.

However, both strategies received an unrealistic change in tilting angles. The tilting angles undergo large changes during a flight [1,22]; some are unexpectedly fast. These requirements are hardly practical; a reasonable tilting angle should be within a small range rather than beyond a period ($2\pi$).

In this research, we prohibit the change of the tilting angles during our flight and analyze each four-tilting-angle combination to find the combinations where feedback linearization is applicable (e.g., the decoupling matrix is invertible). This means that the number of inputs is four in this research. Further, the accepted region for designing the gait is explored in the attitude–altitude stabilization experiment.

In comparison, we simulate the control result in completing the same task equipped with the other controller, which fully utilizes all the inputs (over-actuated) on the same tiltrotor with identical parameters.

The rest of this paper is structured as follows. Section 2 introduces the dynamics of the tiltrotor. Then, a feedback-linearization-based controller is developed in Section 3. The necessary conditions to be an applicable gait are analyzed in Section 4. In the experiment in Section 5, the attitude and the altitude of the tiltrotor are stabilized by the controller designed in Section 3. Section 6 shows the results and the sufficient conditions to be an applicable gait in the region of interest. The conclusions and discussions are addressed in Section 7.

## 2. Dynamics of the Tiltrotor

The tiltrotor is a type of modified quadrotor whose directions of the thrusts are able to change [1–7]. Figure 1 sketches the tiltrotor. As can be seen in Figure 1, the direction of each thrust is in the relevant yellow plane. For details of the kinematics, [1,2,4,5,7] are recommended.

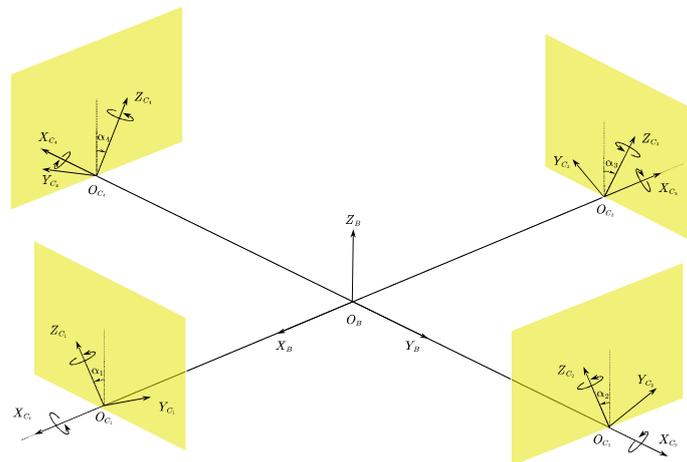

**Figure 1.** The sketch of the tiltrotor.

Among the dynamics of the quadrotor in previous research, the widely accepted [1,2,7,22] model is constructed by separately analyzing each part (body frame, tilting motor frames, and propeller frames) using Newton's Law. The controller is developed based on the simplified dynamics in Equations (1)–(5).

The position $P = [X \quad Y \quad Z]^T$ is ruled by Equation (1).

$$\ddot{p} = \begin{bmatrix} 0 \\ 0 \\ -g \end{bmatrix} + \frac{1}{m} \cdot {}^{W}R \cdot F(\alpha) \cdot \begin{bmatrix} \varpi_1 \cdot |\varpi_1| \\ \varpi_2 \cdot |\varpi_2| \\ \varpi_3 \cdot |\varpi_3| \\ \varpi_4 \cdot |\varpi_4| \end{bmatrix} \triangleq \begin{bmatrix} 0 \\ 0 \\ -g \end{bmatrix} + \frac{1}{m} \cdot {}^{W}R \cdot F(\alpha) \cdot w \tag{1}$$

where $m$ is the total mass. $g$ is the gravitational acceleration. $\varpi_i$, ($i = 1,2,3,4$) is the angular velocity of the propeller ($\varpi_{1,3} < 0$, $\varpi_{2,4} > 0$) with respect to the propeller-fixed frame. ${}^{W}R$ is the rotational matrix [46] from the inertial frame to the body-fixed frame (Equation (2)).

$$^{W}R = \begin{bmatrix} c\theta \cdot c\psi & s\phi \cdot s\theta \cdot c\psi - c\phi \cdot s\psi & c\phi \cdot s\theta \cdot c\psi + s\phi \cdot s\psi \\ c\theta \cdot s\psi & s\phi \cdot s\theta \cdot s\psi + c\phi \cdot c\psi & c\phi \cdot s\theta \cdot s\psi - s\phi \cdot c\psi \\ -s\theta & s\phi \cdot c\theta & c\phi \cdot c\theta \end{bmatrix} \tag{2}$$

where $s\Lambda = \sin (\Lambda)$ and $c\Lambda = \cos (\Lambda)$. $\phi$, $\theta$, and $\psi$ are roll angle, pitch angle, and yaw angle, respectively. Tilting angles $\alpha = [\alpha_1 \quad \alpha_2 \quad \alpha_3 \quad \alpha_4]$. The positive directions of the tilting angles are defined in Figure 1. $F(\alpha)$ is defined in Equation (3).

$$F(\alpha) = \begin{bmatrix} 0 & K_f \cdot s2 & 0 & -K_f \cdot s4 \\ K_f \cdot s1 & 0 & -K_f \cdot s3 & 0 \\ -K_f \cdot c1 & K_f \cdot c2 & -K_f \cdot c3 & K_f \cdot c4 \end{bmatrix} \tag{3}$$

where $si = \sin (\alpha_i)$, $ci = \cos (\alpha_i)$, and ($i = 1,2,3,4$). $K_f$ ($8.048 \times 10^{-6}$) is the coefficient of the thrust.

The angular velocity of the body with respect to its own frame, $\omega_B = [p \quad q \quad r]^T$, is governed by Equation (4).

$$\dot{\omega}_B = I_B^{-1} \cdot \tau(\alpha) \cdot w \tag{4}$$

where $I_B$ is the matrix of moments of inertia. $\tau(\alpha)$ is defined in Equation (5). $K_m$ ($2.423 \times 10^{-7}$) is the coefficient of the drag. $L$ is the length of the arm.

$$\tau(\alpha) = \begin{bmatrix} 0 & L \cdot K_f \cdot c2 - K_m \cdot s2 & 0 & -L \cdot K_f \cdot c4 + K_m \cdot s4 \\ L \cdot K_f \cdot c1 + K_m \cdot s1 & 0 & -L \cdot K_f \cdot c3 - K_m \cdot s3 & 0 \\ L \cdot K_f \cdot s1 - K_m \cdot c1 & -L \cdot K_f \cdot s2 - K_m \cdot c2 & L \cdot K_f \cdot s3 - K_m \cdot c3 & -L \cdot K_f \cdot s4 - K_m \cdot c4 \end{bmatrix} \tag{5}$$

So far, we have determined the dynamics of the tiltrotor. There are several remarks.

Firstly, the tilting angles ($\alpha_1, \alpha_2, \alpha_3, \alpha_4$) are predetermined before controlling. This indicates that they are constant during a flight, e.g., Equation (6) holds.

$$\dot{\alpha}_i \equiv 0, i = 1, 2, 3, 4 \tag{6}$$

In this research, we will find the proper tilting angles to support our flight.

Secondly, the relationship [47–49] between the angular velocity of the body, $\omega_B$, and the attitude rotation matrix (${}^{W}R$) is given in Equation (7).

$$^{W}\dot{R} = {}^{W}R \cdot \hat{\omega}_B \tag{7}$$

where "$\hat{\ }$" is the hat operation to produce the skew matrix.

Our simulator is built based on Equations (1)–(7).

While in the controller design process, we approximate the relationship between the angular velocity of the body, $\omega_B$, and the attitude angle, ($\phi, \theta, \psi$) in Equation (8).

$$\begin{bmatrix} \dot{\phi} \\ \dot{\theta} \\ \dot{\psi} \end{bmatrix} = \omega_B \tag{8}$$

Instead of further exploiting Equation (7), the controller is designed based on Equations (1)–(6) and (8).

The parameters of this tiltrotor are specified as follows: $m = 0.429 \ kg$, $L = 0.1785 \ m$, $g = 9.8 \ N/kg$, and $I_B = \text{diag} ([2.24 \times 10^{-3}, 2.99 \times 10^{-3}, 4.80 \times 10^{-3}]) kg \cdot m^2$.

## 3. Feedback Linearization and Control

The control scenario consists of two sections. Firstly, the nonlinear dynamics are dynamically inverted by feedback linearization. Secondly, the linearized system is stabilized based on a third-order PD controller. The rest of this section introduces these strategies.

### 3.1. Feedback Linearization

The first step in feedback linearization is to select the output. In this research, we choose attitude–altitude in Equation (9) as our output, since the choice of position–yaw may introduce a further singular zone [50,51].

$$\begin{bmatrix} y_1 \\ y_2 \\ y_3 \\ y_4 \end{bmatrix} = \begin{bmatrix} \phi \\ \theta \\ \psi \\ Z \end{bmatrix} \tag{9}$$

Calculating the second derivative of Equation (9) yields Equation (10).

$$\begin{bmatrix} \ddot{y}_1 \\ \ddot{y}_2 \\ \ddot{y}_3 \\ \ddot{y}_4 \end{bmatrix} = \begin{bmatrix} 0 \\ 0 \\ 0 \\ -g \end{bmatrix} + \begin{bmatrix} I_B^{-1} \cdot \tau(\alpha) \\ [0 \quad 0 \quad 1] \cdot \dfrac{K_f}{m} \cdot {}^W R \cdot F(\alpha) \end{bmatrix}^{4\times 4} \cdot w \tag{10}$$

Notice that we have Equation (11) if $\varpi_{1,3} < 0$, $\varpi_{2,4} > 0$.

$$(\varpi_i \cdot |\varpi_i|)' = 2 \cdot \dot{\varpi}_i \cdot |\varpi_i| \tag{11}$$

Differentiating Equation (10) yields Equation (12).

$$\begin{bmatrix} \dddot{y}_1 \\ \dddot{y}_2 \\ \dddot{y}_3 \\ \dddot{y}_4 \end{bmatrix} = \begin{bmatrix} I_B^{-1} \cdot \tau(\alpha) \\ [0 \quad 0 \quad 1] \cdot \dfrac{K_f}{m} \cdot {}^W R \cdot F(\alpha) \cdot 2 \cdot \begin{bmatrix} |\varpi_1| & & & \\ & |\varpi_2| & & \\ & & |\varpi_3| & \\ & & & |\varpi_4| \end{bmatrix} \end{bmatrix}^{4\times 4} \cdot \begin{bmatrix} \dot{\varpi}_1 \\ \dot{\varpi}_2 \\ \dot{\varpi}_3 \\ \dot{\varpi}_4 \end{bmatrix} + [0 \quad 0 \quad 1] \cdot \dfrac{K_f}{m} \cdot {}^W R \cdot \hat{\omega}_B \cdot F(\alpha)$$

$$\cdot w \cdot \begin{bmatrix} 0 \\ 0 \\ 0 \\ 1 \end{bmatrix} \triangleq \bar{\Delta} \cdot \begin{bmatrix} \dot{\varpi}_1 \\ \dot{\varpi}_2 \\ \dot{\varpi}_3 \\ \dot{\varpi}_4 \end{bmatrix} + Ma \tag{12}$$

$\bar{\Delta}$ is called the decoupling matrix [32]. $[\dot{\varpi}_1 \quad \dot{\varpi}_2 \quad \dot{\varpi}_3 \quad \dot{\varpi}_4]^T \triangleq U$ is the new input vector.

From Equation (12), we may receive the decoupled relationship in Equation (13) compatible with the controller design process.

$$\begin{bmatrix} \dot{\varpi}_1 \\ \dot{\varpi}_2 \\ \dot{\varpi}_3 \\ \dot{\varpi}_4 \end{bmatrix} = \bar{\Delta}^{-1} \cdot \left( \begin{bmatrix} \dddot{y}_{1d} \\ \dddot{y}_{2d} \\ \dddot{y}_{3d} \\ \dddot{y}_{4d} \end{bmatrix} - Ma \right) \tag{13}$$

Obviously, the necessary condition for receiving Equation (13) is that the decoupling matrix ($\bar{\Delta}$) is invertible. Section 4 in this paper deepens this discussion.

Once receiving Equation (13), the controller may be applied to this linearized system. In this research, we deploy third-order PD controllers.

### 3.2. Third-Order PD Controllers

We design the following third-order PD controllers in Equations (14) and (15).

$$\begin{bmatrix} \dddot{y}_{1d} \\ \dddot{y}_{2d} \\ \dddot{y}_{3d} \end{bmatrix} = \begin{bmatrix} \dddot{y}_{1r} \\ \dddot{y}_{2r} \\ \dddot{y}_{3r} \end{bmatrix} + K_{P1}{}^{3\times 3} \cdot \left( \begin{bmatrix} \ddot{y}_{1r} \\ \ddot{y}_{2r} \\ \ddot{y}_{3r} \end{bmatrix} - \begin{bmatrix} \ddot{y}_1 \\ \ddot{y}_2 \\ \ddot{y}_3 \end{bmatrix} \right) + K_{P2}{}^{3\times 3} \cdot \left( \begin{bmatrix} \dot{y}_{1r} \\ \dot{y}_{2r} \\ \dot{y}_{3r} \end{bmatrix} - \begin{bmatrix} \dot{y}_1 \\ \dot{y}_2 \\ \dot{y}_3 \end{bmatrix} \right) + K_{P3}{}^{3\times 3} \cdot \left( \begin{bmatrix} y_{1r} \\ y_{2r} \\ y_{3r} \end{bmatrix} - \begin{bmatrix} y_1 \\ y_2 \\ y_3 \end{bmatrix} \right) \tag{14}$$

$$\ddot{y}_{4d} = \ddot{y}_{4r} + K_{PZ_1} \cdot (\dot{y}_{4r} - \dot{y}_4) + K_{PZ_2} \cdot (\dot{y}_{4r} - \dot{y}_4) + K_{PZ_3} \cdot (y_{4r} - y_4) \tag{15}$$

where $K_{Pi}$ ($i = 1,2,3$) is the 3-by-3 diagonal control coefficient matrix. $K_{PZ_i}$ ($i = 1,2,3$) is the control coefficient (scalar). $y_j$ ($j = 1, 2, 3, 4$) is the state. $y_{jr}$ ($j = 1,2,3,4$) is the reference.

The control parameters in this section are specified as follows: $K_{P1} = K_{P2} = K_{P3} = \text{diag}([1,1,1])$, $K_{PZ_1} = 10$, $K_{PZ_2} = 5$, and $K_{PZ_3} = 10$.

## 4. Applicable Gait (Necessary Conditions)

As discussed, the necessary condition for applying this control is that the decoupling matrix ($\bar{\Delta}$) is invertible. In this section, we find the equivalent conditions to receive an invertible decoupling matrix.

Notice the relationship in (16).

$$\bar{\Delta} \sim \begin{bmatrix} \tau(\alpha) \\ [0 \quad 0 \quad 1] \cdot {}^W R \cdot F(\alpha) \end{bmatrix} \tag{16}$$

where "$A \sim B$" represents that Matrix $A$ is equivalent to Matrix $B$. Two matrices are called equivalent if and only if there exist invertible matrices $P$ and $Q$, so that $A = P \cdot B \cdot Q$.

The following propositions and proofs are specifically applicable only to the tiltrotor with our chosen control parameters and the coefficients.

**Proposition 1.** *The decoupling matrix is invertible if and only if Condition (17) holds.*

$$\begin{aligned}
&1.000 \cdot c1 \cdot c2 \cdot c3 \cdot s4 \cdot s\theta - 1.000 \cdot c1 \cdot s2 \cdot c3 \cdot c4 \cdot s\theta - 2.880 \cdot c1 \cdot c2 \cdot s3 \\
&\cdot s4 \cdot s\theta + 2.880 \cdot c1 \cdot s2 \cdot s3 \cdot c4 \cdot s\theta - 2.880 \cdot s1 \cdot c2 \cdot c3 \cdot s4 \\
&\cdot s\theta + 2.880 \cdot s1 \cdot s2 \cdot c3 \cdot c4 \cdot s\theta - 1.000 \cdot s1 \cdot c2 \cdot s3 \cdot s4 \cdot s\theta \\
&+ 1.000 \cdot s1 \cdot s2 \cdot s3 \cdot c4 \cdot s\theta + 4.000 \cdot c1 \cdot c2 \cdot c3 \cdot c4 \cdot c\phi \cdot s\theta \\
&+ 5.592 \cdot c1 \cdot c2 \cdot c3 \cdot s4 \cdot c\phi \cdot c\theta - 5.592 \cdot c1 \cdot c2 \cdot s3 \cdot c4 \\
&\cdot c\phi \cdot c\theta + 5.592 \cdot c1 \cdot s2 \cdot c3 \cdot c4 \cdot c\phi \cdot c\theta - 5.592 \cdot s1 \cdot c2 \\
&\cdot c3 \cdot c4 \cdot c\phi \cdot c\theta + 1.000 \cdot c1 \cdot c2 \cdot s3 \cdot c4 \cdot s\phi \cdot c\theta + 0.9716 \\
&\cdot c1 \cdot c2 \cdot s3 \cdot s4 \cdot c\phi \cdot c\theta - 2.000 \cdot c1 \cdot s2 \cdot c3 \cdot s4 \cdot c\phi \cdot c\theta \\
&+ 0.9716 \cdot c1 \cdot s2 \cdot s3 \cdot c4 \cdot c\phi \cdot c\theta - 1.000 \cdot s1 \cdot c2 \cdot c3 \cdot c4 \\
&\cdot s\phi \cdot c\theta + 0.9716 \cdot s1 \cdot c2 \cdot c3 \cdot s4 \cdot c\phi \cdot c\theta - 2.000 \cdot s1 \cdot c2 \\
&\cdot s3 \cdot c4 \cdot c\phi \cdot c\theta + 0.9716 \cdot s1 \cdot s2 \cdot c3 \cdot c4 \cdot c\phi \cdot c\theta + 2.880 \\
&\cdot c1 \cdot c2 \cdot s3 \cdot s4 \cdot s\phi \cdot c\theta + 2.880 \cdot c1 \cdot s2 \cdot s3 \cdot c4 \cdot s\phi \cdot c\theta \\
&- 0.1687 \cdot c1 \cdot s2 \cdot s3 \cdot s4 \cdot c\phi \cdot c\theta - 2.880 \cdot s1 \cdot c2 \cdot c3 \cdot s4 \\
&\cdot s\phi \cdot c\theta + 0.1687 \cdot s1 \cdot c2 \cdot s3 \cdot s4 \cdot c\phi \cdot c\theta - 2.880 \cdot s1 \\
&\cdot c3 \cdot c4 \cdot s\phi \cdot c\theta - 0.1687 \cdot s1 \cdot s2 \cdot c3 \cdot s4 \cdot c\phi \cdot c\theta + 0.1687 \\
&\cdot s1 \cdot s2 \cdot s3 \cdot c4 \cdot c\phi \cdot c\theta - 1.000 \cdot c1 \cdot s2 \cdot s3 \cdot s4 \cdot s\phi \cdot c\theta \\
&+ 1.000 \cdot s1 \cdot s2 \cdot c3 \cdot s4 \cdot s\phi \cdot c\theta \neq 0
\end{aligned} \tag{17}$$

**Proof of Proposition 1.** *Expanding the second matrix in (16) yields (18).*

$$\bar{\Delta} \sim \begin{bmatrix} 0 & L \cdot K_f \cdot c2 - K_m \cdot s2 & 0 & -L \cdot K_f \cdot c4 + K_m \cdot s4 \\ L \cdot K_f \cdot c1 + K_m \cdot s1 & 0 & -L \cdot K_f \cdot c3 - K_m \cdot s3 & 0 \\ L \cdot K_f \cdot s1 - K_m \cdot c1 & -L \cdot K_f \cdot s2 - K_m \cdot c2 & L \cdot K_f \cdot s3 - K_m \cdot c3 & -L \cdot K_f \cdot s4 - K_m \cdot c4 \\ c\theta \cdot s\phi \cdot s1 - c\theta \cdot c\phi \cdot c1 & -s\theta \cdot s2 + c\theta \cdot c\phi \cdot c2 & -c\theta \cdot s\phi \cdot s3 - c\theta \cdot c\phi \cdot c3 & s\theta \cdot s4 + c\theta \cdot c\phi \cdot c4 \end{bmatrix} \tag{18}$$

*Calculating the determinant of the second matrix in (18) yields Condition (17).* □

**Proposition 2.** *When the roll angle and pitch angle of the tiltrotor are close to zero, the decoupling matrix is invertible if and only if Condition (19) holds.*

$$
\begin{aligned}
4.000 &\cdot c1 \cdot c2 \cdot c3 \cdot c4 + 5.592 \\
&\cdot (+c1 \cdot c2 \cdot c3 \cdot s4 - c1 \cdot c2 \cdot s3 \cdot c4 + c1 \cdot s2 \cdot c3 \cdot c4 - s1 \cdot c2 \cdot c3 \cdot c4) \\
&+ 0.9716 \\
&\cdot (+c1 \cdot c2 \cdot s3 \cdot s4 + c1 \cdot s2 \cdot s3 \cdot c4 + s1 \cdot c2 \cdot c3 \cdot s4 + s1 \cdot s2 \cdot c3 \cdot c4) \\
&+ 2.000 \cdot (-c1 \cdot s2 \cdot c3 \cdot s4 - s1 \cdot c2 \cdot s3 \cdot c4) + 0.1687 \\
&\cdot (-c1 \cdot s2 \cdot s3 \cdot s4 + s1 \cdot c2 \cdot s3 \cdot s4 - s1 \cdot s2 \cdot c3 \cdot s4 + s1 \cdot s2 \cdot s3 \cdot c4) \neq 0
\end{aligned}
\tag{19}
$$

**Proof of Proposition 2.** *Make the assumptions in Equations (20) and (21).*

$$\theta = 0 \tag{20}$$

$$\phi = 0 \tag{21}$$

*Substituting Equations (20) and (21) into Condition (17) yields Condition (19).* □

**Remark 1.** *One may believe that Equation (18) would be an applicable gait-selecting zone where any combination of $(\alpha_1, \alpha_2, \alpha_3, \alpha_4)$ satisfying Proposition 1 would lead to a stable result. Unfortunately, it is not always true, since the proved Proposition guarantees the invertibility of the decoupling matrix only. Hitting the non-negative angular velocity bound for $\varpi_i$ can result in instability. Obviously, Proposition 1 or 2 does not rule out this situation. That is the reason we call the Propositions the "necessary" conditions for an applicable gait. The sufficient condition within the range of interest is found in the experiment in Sections 5 and 6.*

**Remark 2.** *Notice that Condition (17) not only restricts the gait $(\alpha_1, \alpha_2, \alpha_3, \alpha_4)$ but also rules out some attitudes $(\phi, \theta)$; a specific gait can be driven to violate Condition (17) while steering to some specific attitudes. Actually, the similar attitude-based condition causes the failure to hold the decoupling matrix invertible in [50], hindering further applications in feedback linearization (position–yaw output) before modification. This is because roll and pitch are not directly controlled based on the position–yaw output choice. However, the adverse effect of this property is weakened in our research. The output choice (attitude–altitude) lets the tiltrotor directly steer the attitude. Thus, roll angle and pitch angle can be relatively arbitrarily assigned. Steering the attitude away from the region violating Condition (17) is consequently possible by this controller.*

## 5. Attitude–Altitude Stabilization Test

The conditions in Section 4 are the necessary conditions for the applicable gait. The sufficient conditions are explored empirically in this section.

To reduce the complexity, the preliminary step is to restrict the gait to some extent and find the interested gait region. The next step is to determine the exploring direction of the candidate gaits. The final step is to conduct the experiment for each gait along the exploring direction.

### 5.1. Restricted Gait Region

The experiment was conducted near zero roll angle and pitch angle ($\phi = 0, \theta = 0$). Thus, the further analysis is based on Condition (19) rather than on Condition (17).

Although Condition (19) includes no information about roll angle and pitch angle, there are still four parameters in the gait $(\alpha_1, \alpha_2, \alpha_3, \alpha_4)$ to be determined, which is complicated for further discussions. In this consideration, we simplify the gait beforehand.

Instead of exploring the entire space of $(\alpha_1, \alpha_2, \alpha_3, \alpha_4)$, we explore four gaits with one of the following restrictions in each gait: $\alpha_1 = \alpha_3$, $\alpha_1 = \frac{1}{2} \cdot \alpha_3$, $\alpha_1 = -\alpha_3$, and $\alpha_1 = -\frac{1}{2} \cdot \alpha_3$.

#### 5.1.1. Case 1 (Equal)

This section demonstrates the result for some cases satisfying $\alpha_1 = \alpha_3$. Specifically, $\alpha_1 = \alpha_3 = -0.15, -0.075, 0, 0.075, 0.15$. Figure 2 plots the left side of Condition (19). The result is five surfaces about $(\alpha_2, \alpha_4)$. Intercepting the surfaces in Figure 2 by the zero plane (Determinant $= 0$) yields Figure 3. Figure 3 plots the $(\alpha_2, \alpha_4)$ violating Condition (19).

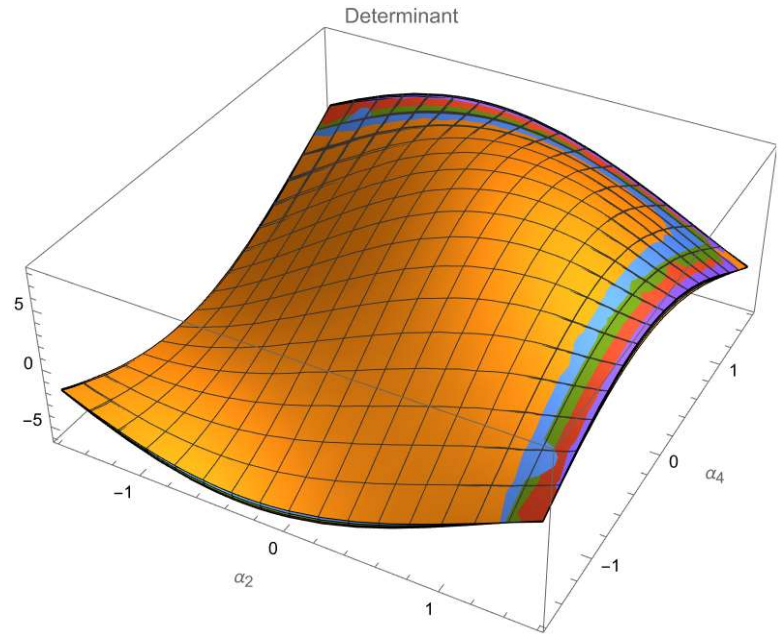

**Figure 2.** Surfaces of the determinant ($\alpha_1 = \alpha_3$).

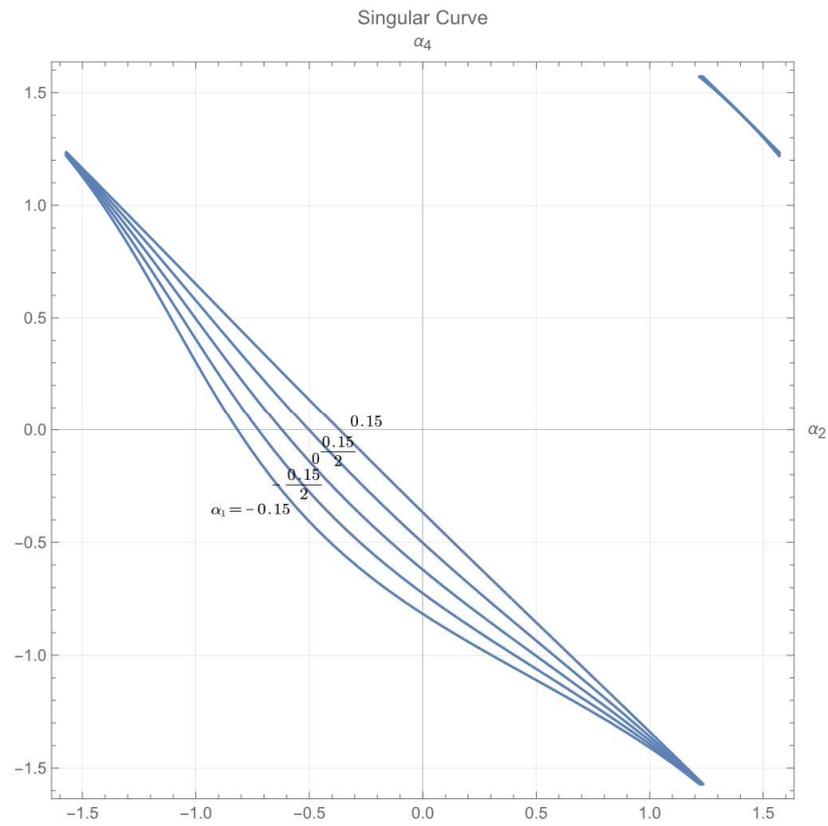

**Figure 3.** $(\alpha_2, \alpha_4)$ violating Condition (19) ($\alpha_1 = \alpha_3$).

### 5.1.2. Case 2 (Half)

This section demonstrates the result for some cases satisfying $\alpha_1 = \frac{1}{2} \cdot \alpha_3$. Specifically, $\alpha_1 = \frac{1}{2} \cdot \alpha_3 = -0.2, -0.1, 0, 0.1, 0.2$, Figure 4 plots the left side of Condition (19). The result is five surfaces about $(\alpha_2, \alpha_4)$. Intercepting the surfaces in Figure 4 by the zero plane (Determinant $= 0$) yields Figure 5. Figure 5 plots the $(\alpha_2, \alpha_4)$ violating Condition (19).

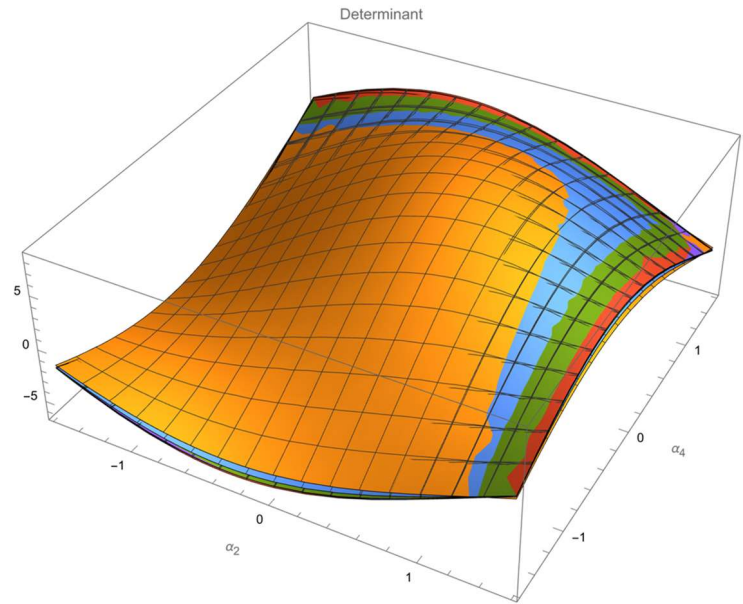

**Figure 4.** Surfaces of the determinant ($\alpha_1 = \frac{1}{2} \cdot \alpha_3$).

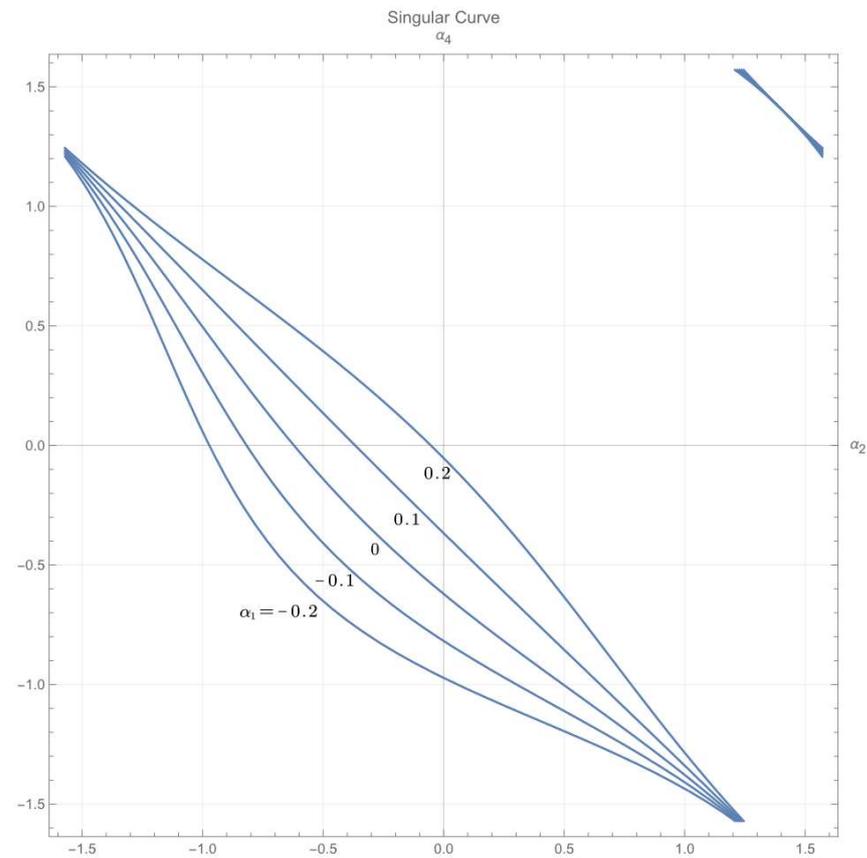

**Figure 5.** $(\alpha_2, \alpha_4)$ violating Condition (19) ($\alpha_1 = \frac{1}{2} \cdot \alpha_3$).

### 5.1.3. Case 3 (Negative)

This section demonstrates the result for some cases satisfying $\alpha_1 = -\alpha_3$. Specifically, $\alpha_1 = -\alpha_3 = -1.4, -0.7, 0, 0.7, 1.4$, Figure 6 plots the left side of Condition (19). The result is five surfaces about $(\alpha_2, \alpha_4)$. Intercepting the surfaces in Figure 6 by the zero plane (Determinant $= 0$) yields Figure 7. Figure 7 plots the $(\alpha_2, \alpha_4)$ violating Condition (19).

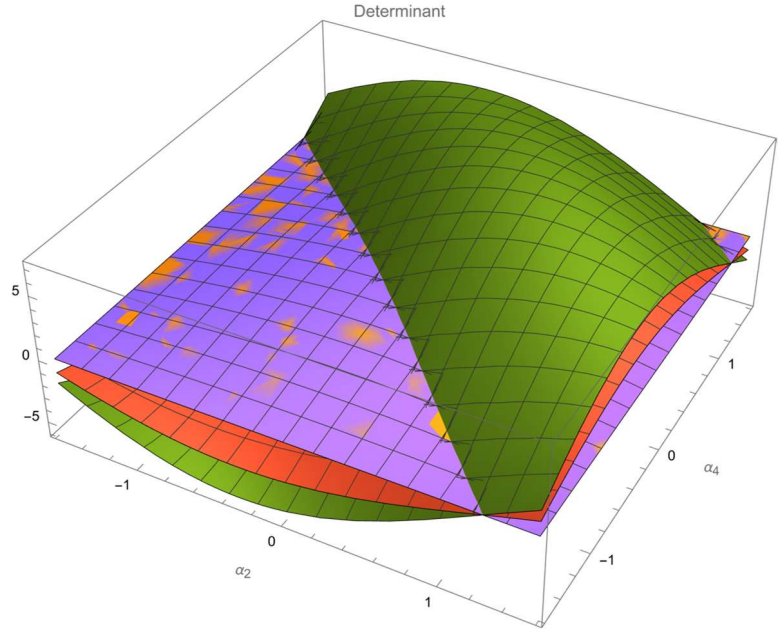

**Figure 6.** Surfaces of the determinant ($\alpha_1 = -\alpha_3$).

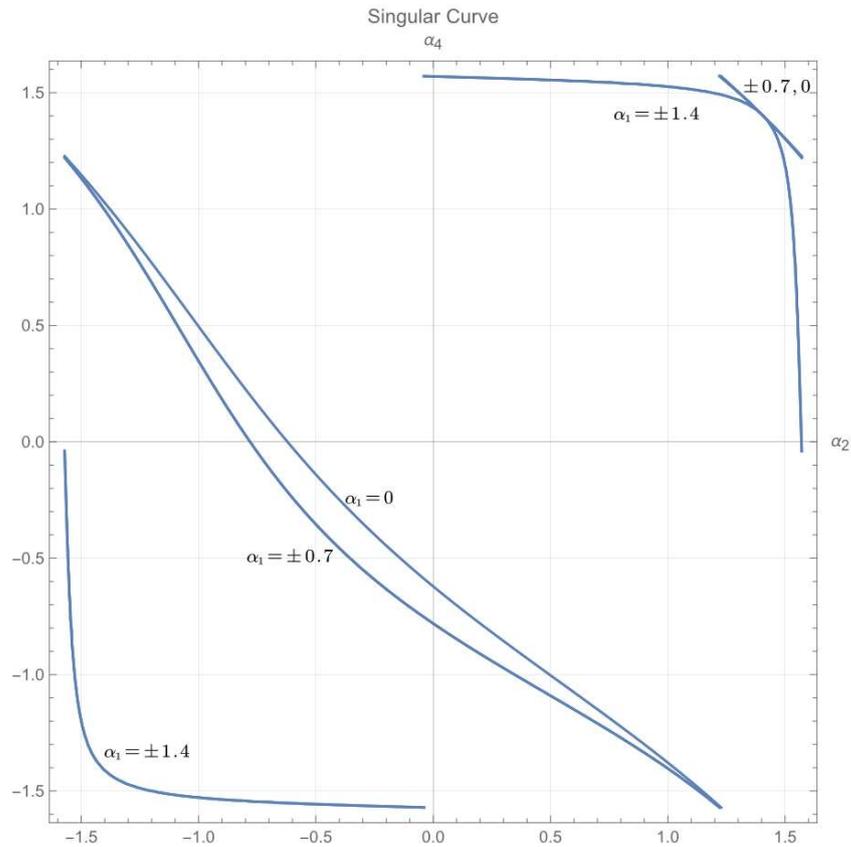

**Figure 7.** $(\alpha_2, \alpha_4)$ violating Condition (19) ($\alpha_1 = -\alpha_3$).

### 5.1.4. Case 4 (Negative Half)

This section demonstrates the result for some cases satisfying $\alpha_1 = -\frac{1}{2} \cdot \alpha_3$. Specifically, $\alpha_1 = -\frac{1}{2} \cdot \alpha_3 = -0.3, -0.15, 0, 0.15, 0.3$, Figure 8 plots the left side of Condition (19). The result is five surfaces about $(\alpha_2, \alpha_4)$. Intercepting the surfaces in Figure 8 by the zero plane (Determinant $= 0$) yields Figure 9. Figure 9 plots the $(\alpha_2, \alpha_4)$ violating Condition (19).

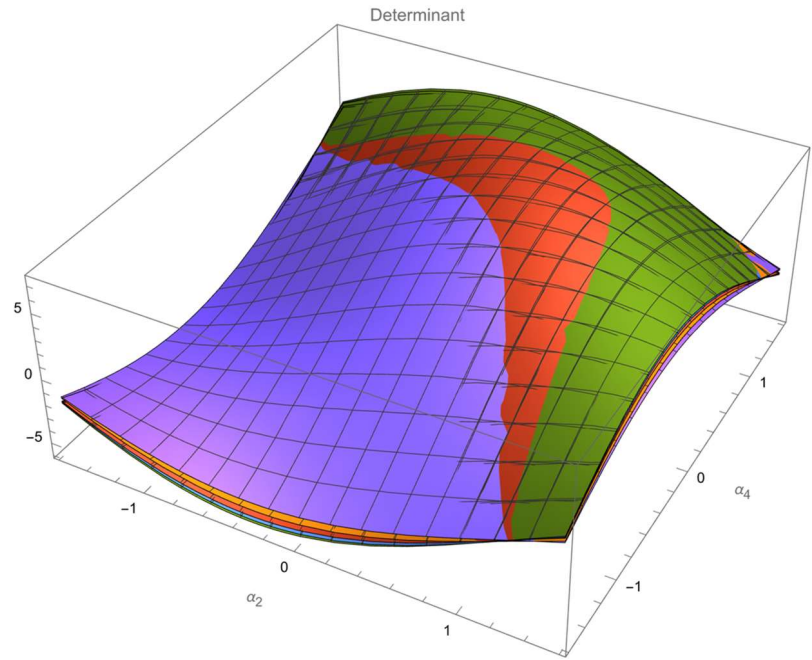

**Figure 8.** Surfaces of the determinant ($\alpha_1 = -\frac{1}{2} \cdot \alpha_3$).

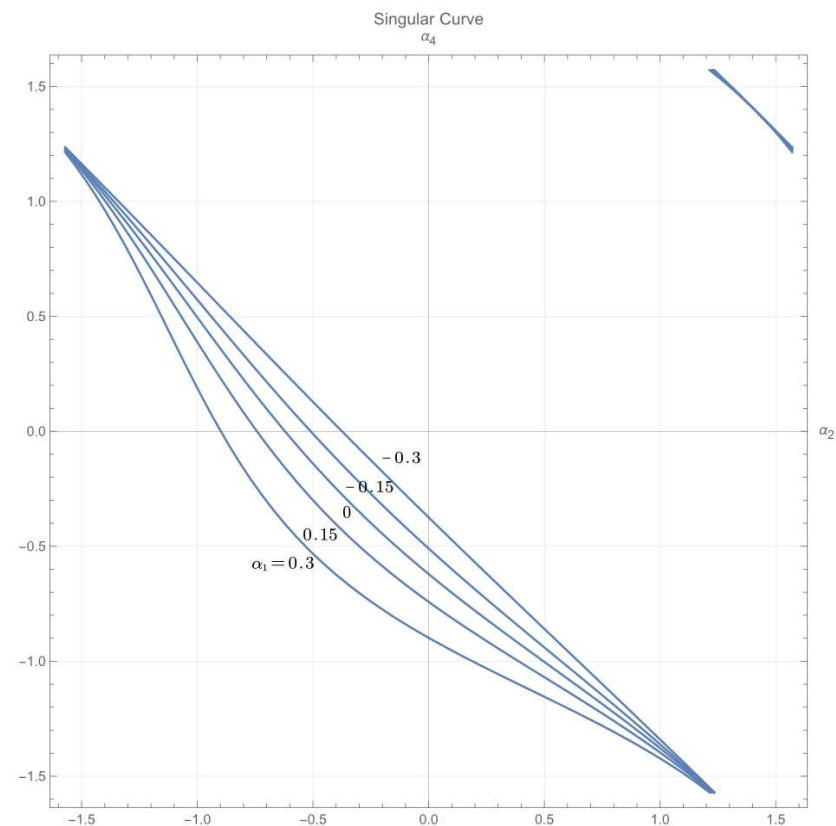

**Figure 9.** $(\alpha_2, \alpha_4)$ violating Condition (19) ($\alpha_1 = -\frac{1}{2} \cdot \alpha_3$).

*5.2. Interested Gait Region and Direction of Exploration*

Notice that choosing the gait of $(\alpha_2,\alpha_4)$ on the relevant curve in Figures 3, 5, 7, and 9 is strictly prohibited. Falling on the relevant curve indicates the violation of Condition (19). It is also worth mentioning that the tiltrotor degrades to the conventional quadrotor for the gait satisfying $(\alpha_1,\alpha_2,\alpha_3,\alpha_4) = (0,0,0,0)$. Our gait analysis includes this special case.

Another point worth mentioning is the continuous requirement in the gait switch. For example, switching from $(\alpha_2,\alpha_4) = (1,1)$ to $(\alpha_2,\alpha_4) = (-1,-1)$ is determined to violate Condition (19) for any case in Figures 3, 5, 7, and 9. This is because the relevant switching process will cross the curve, which should be prohibited.

With these concerns, we only choose part of the region containing $(\alpha_2,\alpha_4) = (0,0)$ as the gait region of our interest.

### 5.2.1. Interested Gait Region

It can be concluded from the green region in Figures 10–13 that any $(\alpha_2,\alpha_4)$ located within or on the edge of the triangular zone defined in (22) in each case in Section 5.1 will not violate Condition (19).

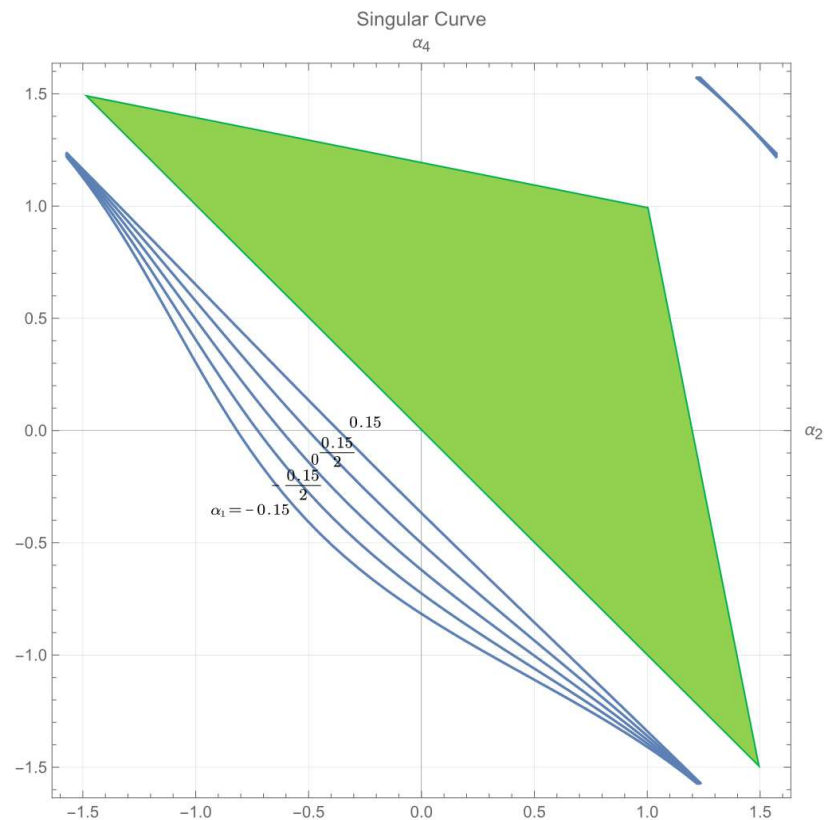

**Figure 10.** The interested invertible triangular region ($\alpha_1 = \alpha_3$).

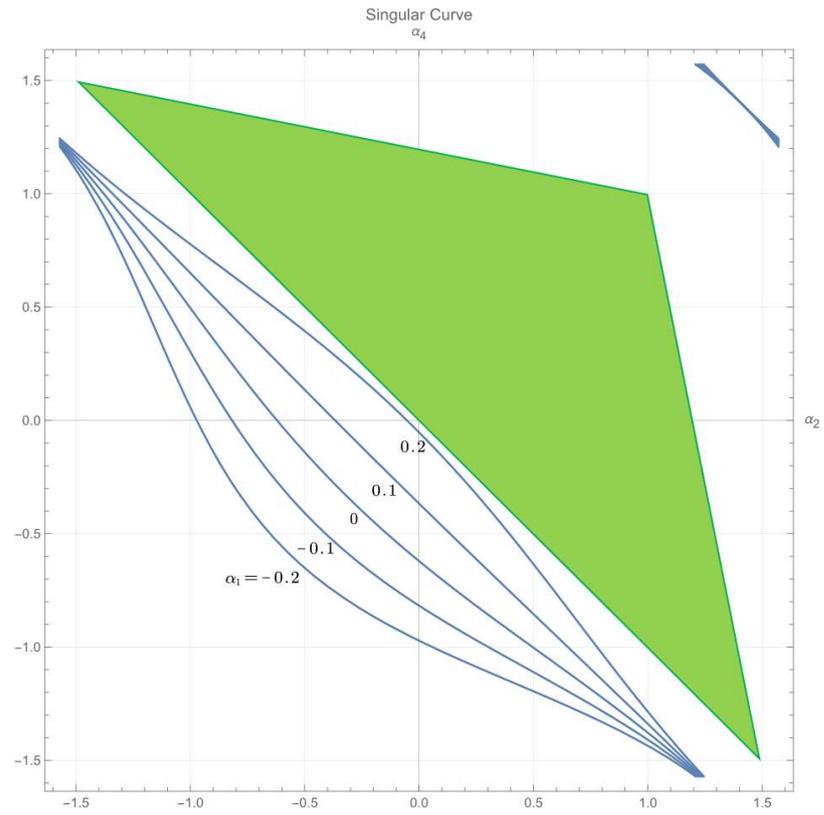

**Figure 11.** The interested invertible triangular region ($\alpha_1 = \frac{1}{2} \cdot \alpha_3$).

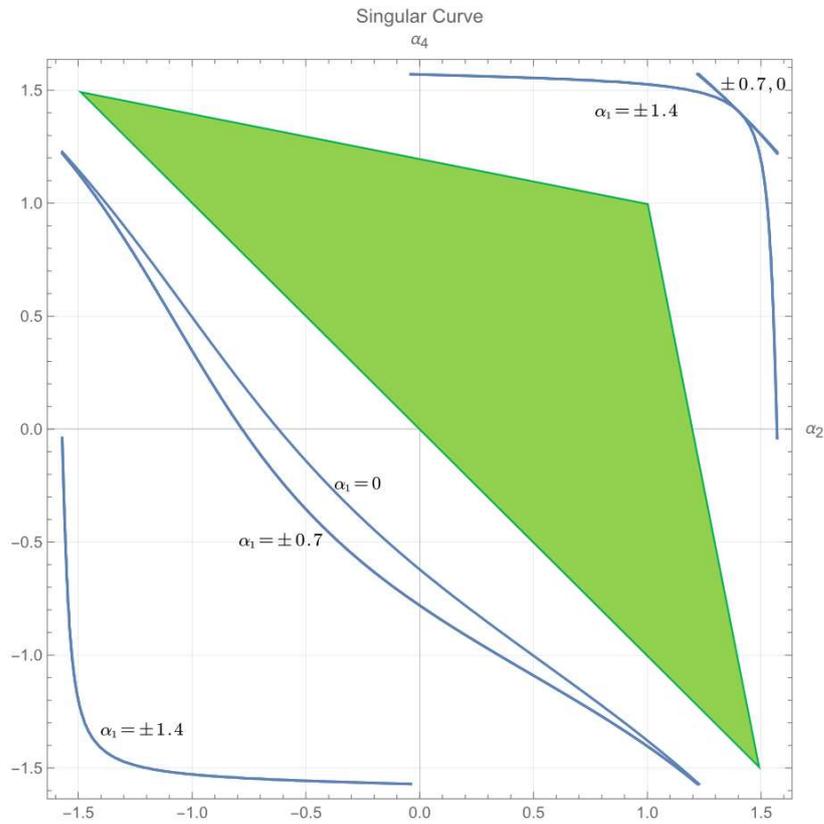

**Figure 12.** The interested invertible triangular region ($\alpha_1 = -\alpha_3$).

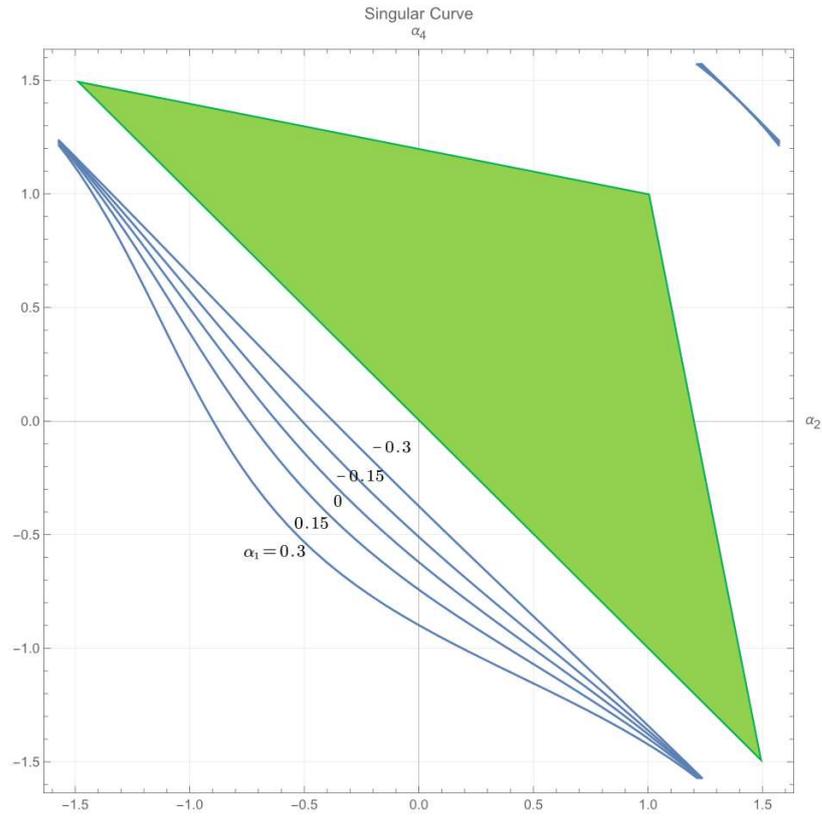

**Figure 13.** The interested invertible triangular region ($\alpha_1 = -\frac{1}{2} \cdot \alpha_3$).

$$\text{Region within or on } \Delta MUV \text{ governed by } U(-1.5,1.5), V(1.5,-1.5), M(1,1) \tag{22}$$

The rest of the paper focuses on the gait ($\alpha_1, \alpha_2, \alpha_3, \alpha_4$) satisfying (22). These gaits naturally satisfy the necessary condition in Proposition 2. We design the direction of exploration in Section 5.2.2 for the experiment to identify the sufficient condition to be stable within the zone in (22).

### 5.2.2. Direction of Exploration

Completing exploring the entire space defined in (22) is not necessary or achievable. Thus, we explore parts of this region based on the definition of directions of exploration in (23)–(25).

$$\alpha_4 = -\alpha_2, \alpha_2 \in \left[-\frac{3}{2}, 0\right] \tag{23}$$

$$\alpha_4 = -\alpha_2, \alpha_2 \in \left[0, \frac{3}{2}\right] \tag{24}$$

$$\alpha_4 = \alpha_2, \alpha_2 \in [0, 1] \tag{25}$$

The exploration along each direction in (23)–(25) starts from $|\alpha_2| = 0$. The exploration ends at a critical $\alpha_{2M}$ defined in (26).

$$\forall \alpha_{2C} \; satisfying \; |\alpha_2| \leqslant |\alpha_{2C}|, (\alpha_1, \alpha_2, \alpha_3, \alpha_4) \Rightarrow stable$$

$$\exists \alpha_2 \; satisfying \; |\alpha_2| > |\alpha_{2C}|, (\alpha_1, \alpha_2, \alpha_3, \alpha_4) \Rightarrow unstable \tag{26}$$

$$\alpha_{2M} = \text{sign}(\alpha_2) \cdot \max(|\alpha_{2C}|)$$

where $\alpha_1$ and $\alpha_3$ are predetermined. $\alpha_4$ is based on the relevant direction in (23)–(25).

The expected output is the three gaits ($\alpha_1, \alpha_2, \alpha_3, \alpha_4$) corresponding to the three $\alpha_{2M}$ along the directions in (23)–(25).

### 5.3. Attitude–Altitude Control

Feedback linearization relies on the choice of output. In this research, the selected output is the attitude–altitude vector; we focus on controlling the attitude and altitude only. Note that this control scheme can also be used to reach a desired position for a conventional quadrotor [52], which can be regarded as the special case where $(\alpha_1, \alpha_2, \alpha_3, \alpha_4) = (0,0,0,0)$ in our tiltrotor. However, the position control for a tiltrotor is not the primary interest of this study.

In this experiment, the tiltrotor is expected to conduct an attitude–altitude self-adjusting performance.

The initial attitude angles of the tiltrotor are assigned as: $\phi_i = 0$, $\theta_i = 0$, and $\phi_i = 0$. The initial angular velocity vector of the tiltrotor with respect to the body-fixed frame is $\omega_B = [0,0,0]^T$. The initial position vector is $[0,0,0]^T$. The initial velocity vector is $[0,0,0]^T$.

Since the input vector is the derivative of each angular velocity of the propeller $[\ddot{\varpi}_1, \ddot{\varpi}_2, \ddot{\varpi}_3, \ddot{\varpi}_4]^T$, assigning the initial velocity for each propeller is necessary. The absolute value of each initial angular velocity of is 300. Notice that these angular velocities are not sufficient to compensate for the effect of gravity, even for the case $(\alpha_1, \alpha_2, \alpha_3, \alpha_4) = (0,0,0,0)$.

The reference is a 4-dimension attitude–altitude vector $[\phi_r, \theta_r, \psi_r, Z_r]^T$. To maintain zero attitude and zero height, the designed reference is $[0,0,0,0]^T$. Based on the control parameters set in Section 3 and each gait designed in Section 5, we recorded the gaits leading to stable results.

## 6. Results

This section presents the results of the experiments in the previous section. Section 6.1 shows the history of the attitude, altitude, and the angular velocities during the flight of two typical gaits. Section 6.2 displays the admissible gaits. They were the stable results of the work explained in Section 5.2.2. Section 6.3 presents the result of completing the same task with another control strategy, which utilized all the inputs (over-actuated) on the same tiltrotor with identical parameters.

### 6.1. Flight History

Section 6.1 presents the results from two gaits. Gait 1: $(\alpha_1, \alpha_2, \alpha_3, \alpha_4) = (-0.1, 0.1, -0.2, 0.1)$. Gait 2: $(\alpha_1, \alpha_2, \alpha_3, \alpha_4) = (-0.15, -0.1, 0.3, -0.1)$.

#### 6.1.1. Gait 1

For the gait $(\alpha_1, \alpha_2, \alpha_3, \alpha_4) = (-0.1, 0.1, -0.2, 0.1)$, the attitude history, altitude history, and the angular velocities history of the propellers are plotted in Figures 14–16, respectively.

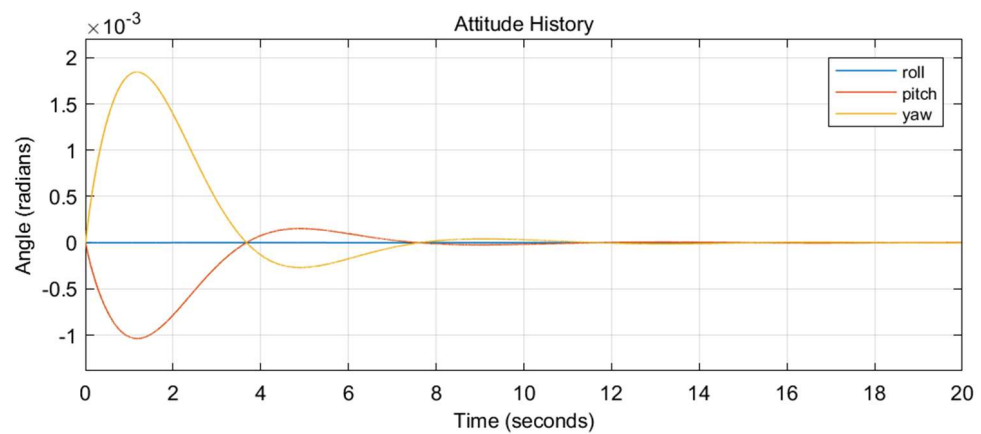

**Figure 14.** Attitude history of Gait 1.

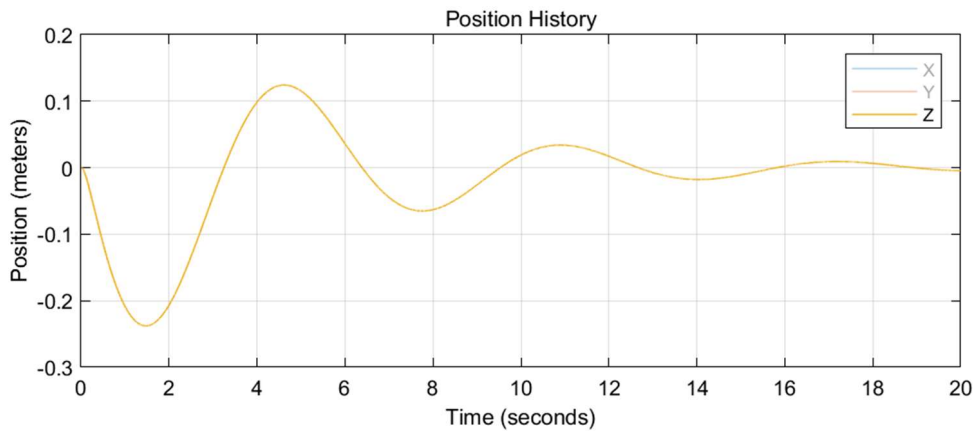

**Figure 15.** Altitude history of Gait 1.

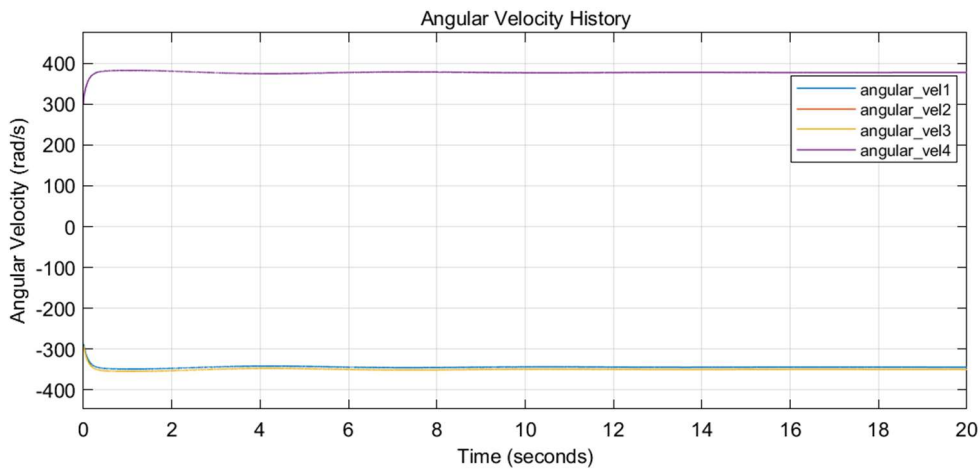

**Figure 16.** Angular velocity history of Gait 1.

Notice that none of the angular velocities touched 0, which guarantees the invertibility of the decoupling matrix in (12).

### 6.1.2. Gait 2

For the gait $(\alpha_1, \alpha_2, \alpha_3, \alpha_4) = (-0.15, -0.1, 0.3, -0.1)$, the attitude history, altitude history, and the angular velocities history of the propellers are plotted in Figures 17–19, respectively.

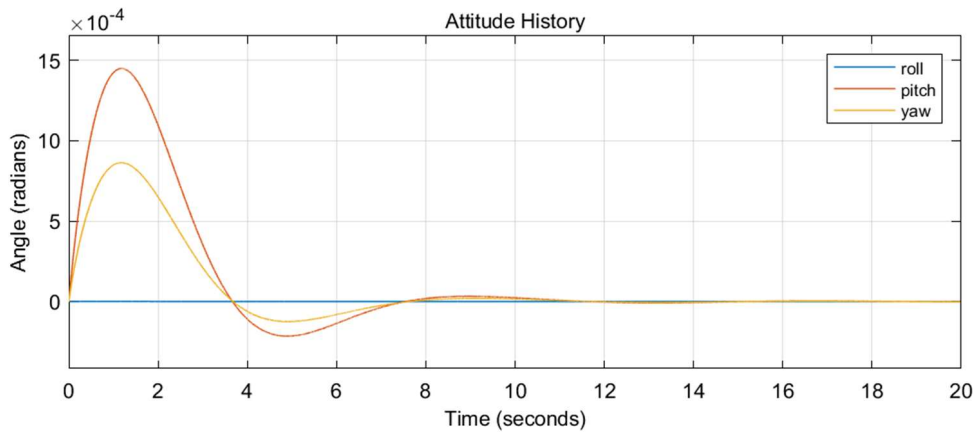

**Figure 17.** Attitude history of Gait 2.

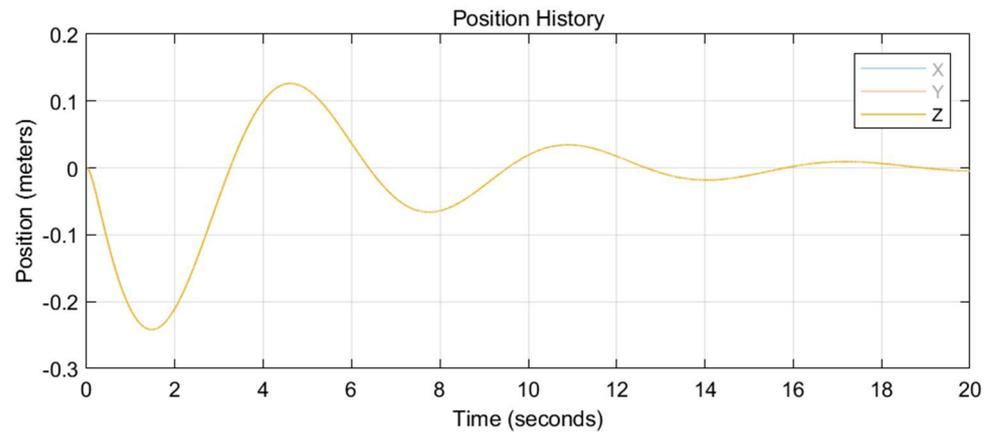

**Figure 18.** Altitude history of Gait 2.

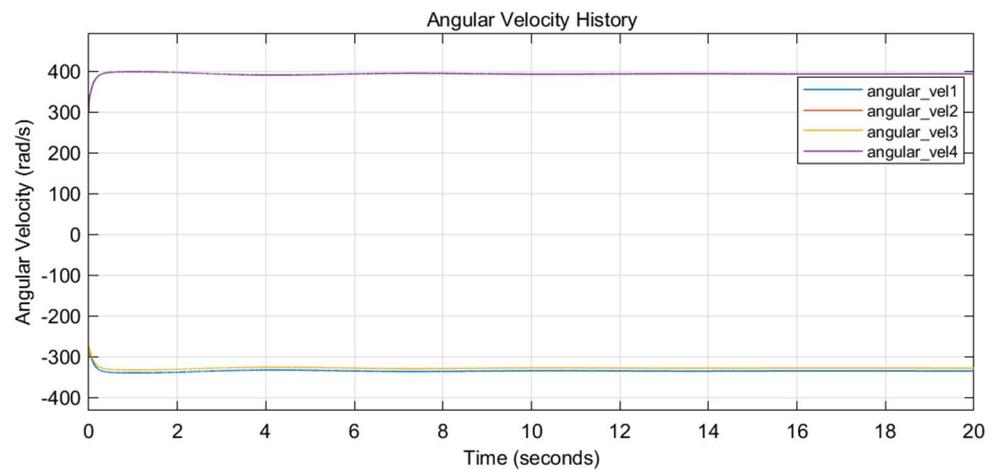

**Figure 19.** Angular velocity history of Gait 2.

Similarly, none of the angular velocities touched 0.

### 6.2. Applicable Gait (Sufficient Conditions)

This section displays the found applicable gaits based on the method in (26) for each gait from Section 5.1.1 to Section 5.1.4 (with five different $(\alpha_1,\alpha_3)$ pairs in each section).

Surprisingly, the resulting applicable $(\alpha_2,\alpha_4)$ leading to stable results in each $(\alpha_1,\alpha_3)$ in 20 (5 × 4) restrictions were highly similar. Except one gait $((\alpha_1,\alpha_3) = (0.2,0.4))$, which led to instability, the regions of the applicable $(\alpha_2,\alpha_4)$ for the remaining 19 cases contained the same triangular region governed by $(-1.3,1.3)$, $(1.3,-1.3)$, and $(1,1)$.

The specific gaits leading to stable control results for different $(\alpha_1,\alpha_3)$ are detailed in Figures 20–23.

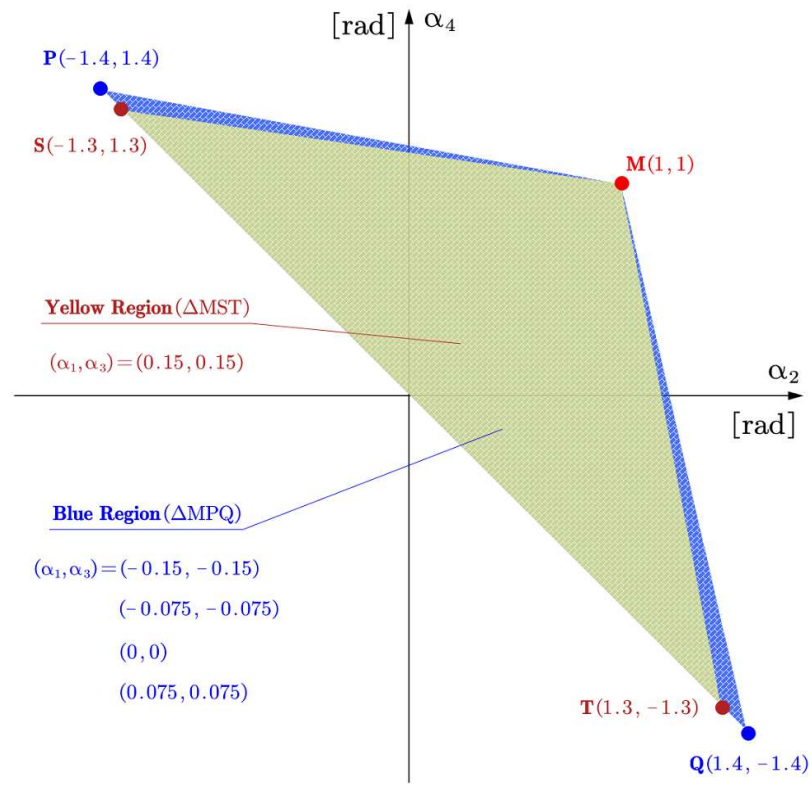

**Figure 20.** Admissible $(\alpha_2, \alpha_4)$ when $\alpha_1 = \alpha_3$.

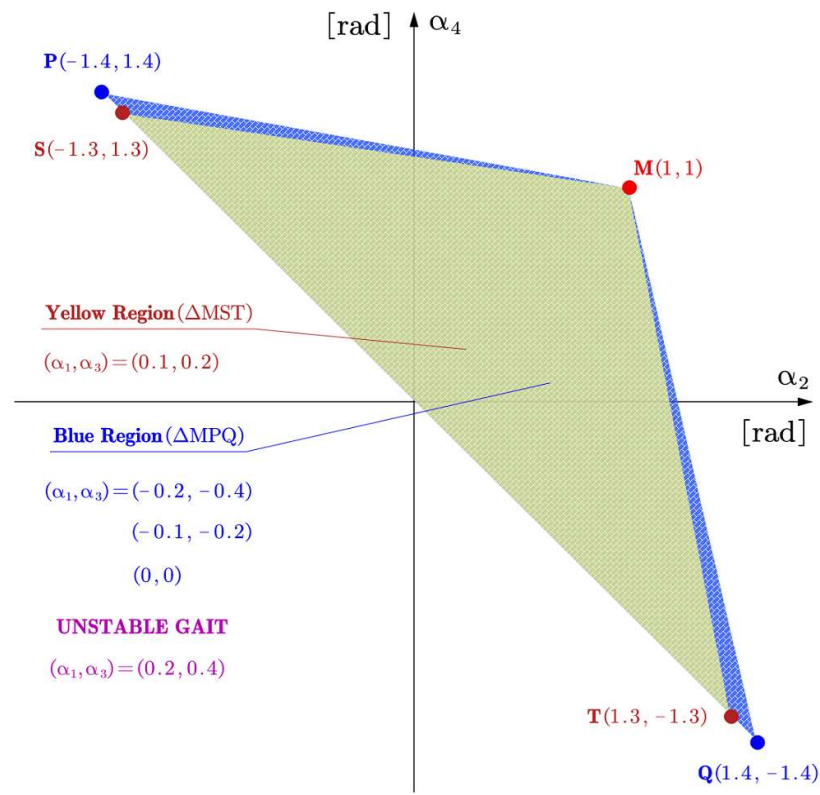

**Figure 21.** Admissible $(\alpha_2, \alpha_4)$ when $\alpha_1 = \frac{1}{2}\alpha_3$.

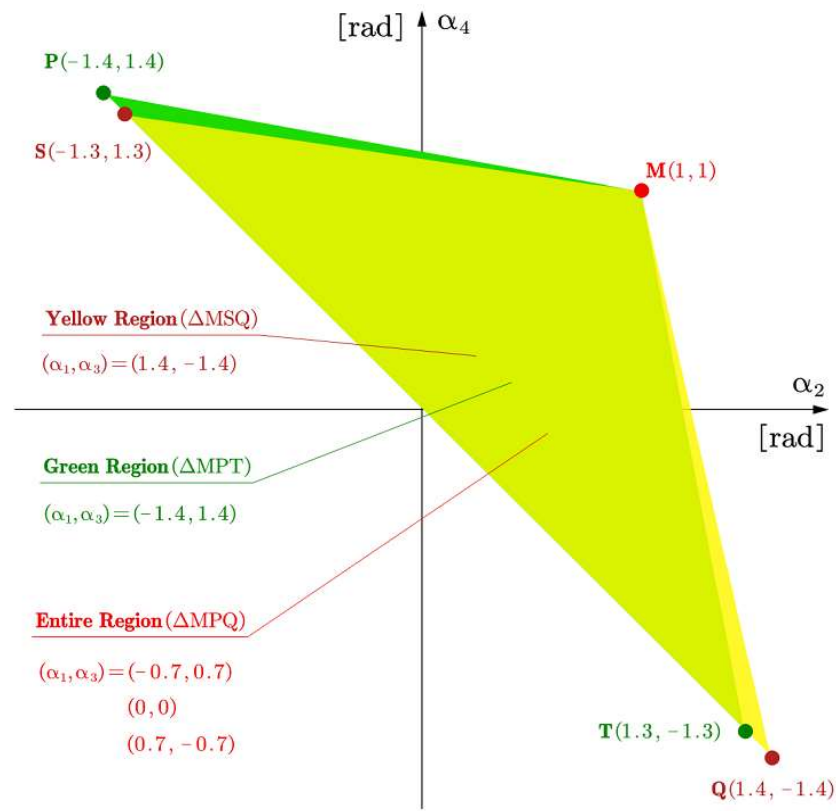

**Figure 22.** Admissible $(\alpha_2, \alpha_4)$ when $\alpha_1 = -\alpha_3$.

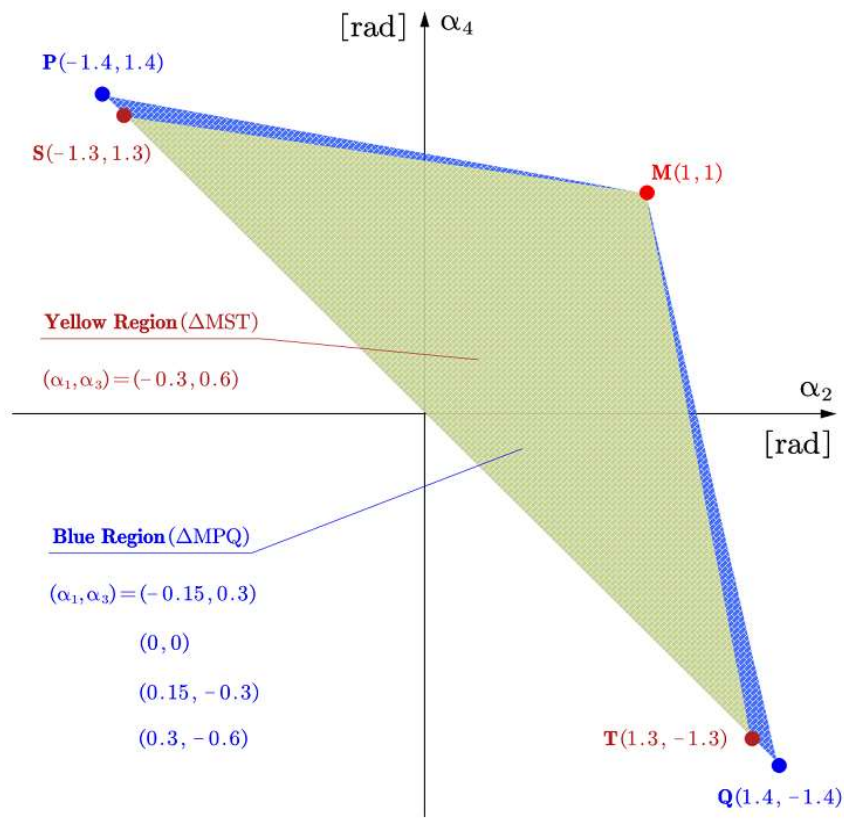

**Figure 23.** Admissible $(\alpha_2, \alpha_4)$ when $\alpha_1 = -\frac{1}{2}\alpha_3$.

*6.3. Over-Actuated Control*

This section presents the control result by an over-actuated controller with eight control inputs in completing the same task with identical parameters and initial conditions. The attitude history, altitude history, the angular velocities history of the propellers, and the tilting angles history are plotted in Figures 24–27, respectively.

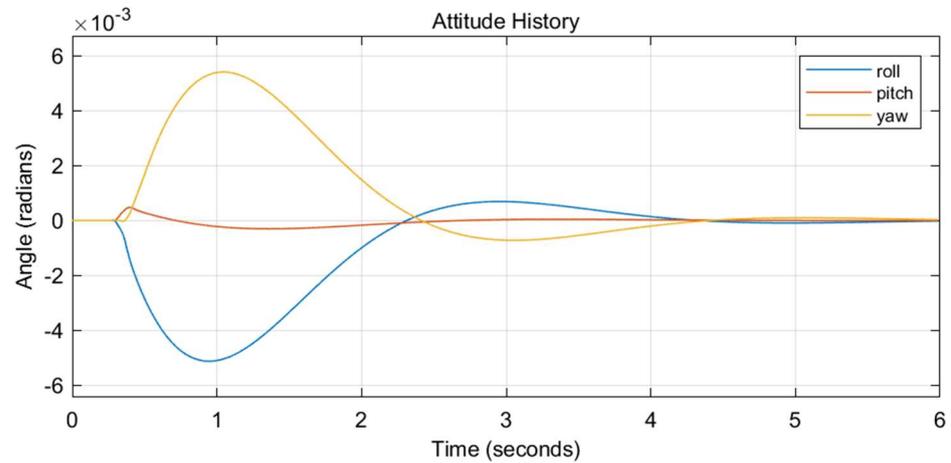

**Figure 24.** Attitude history of over-actuated control.

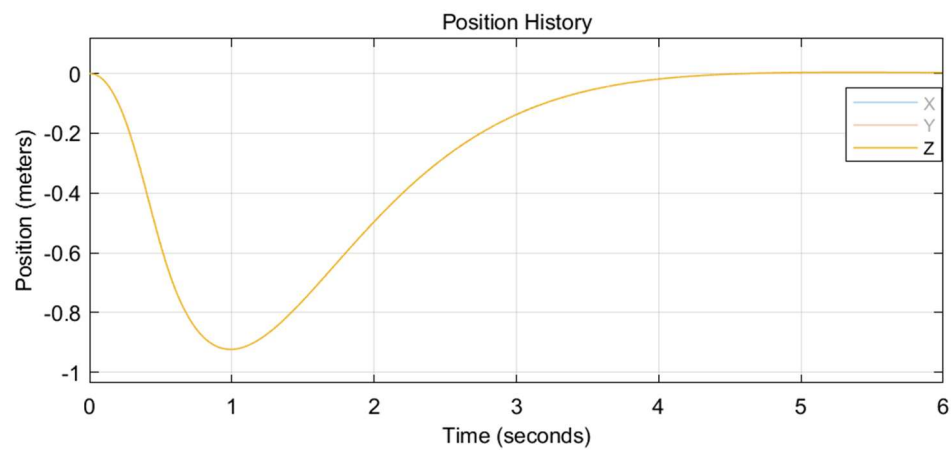

**Figure 25.** Attitude history of over-actuated control.

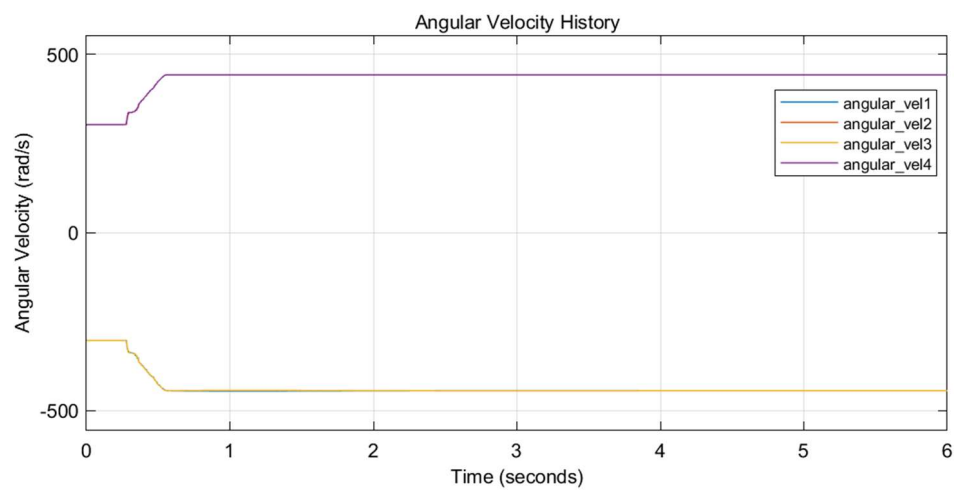

**Figure 26.** Angular velocity history of over-actuated control.

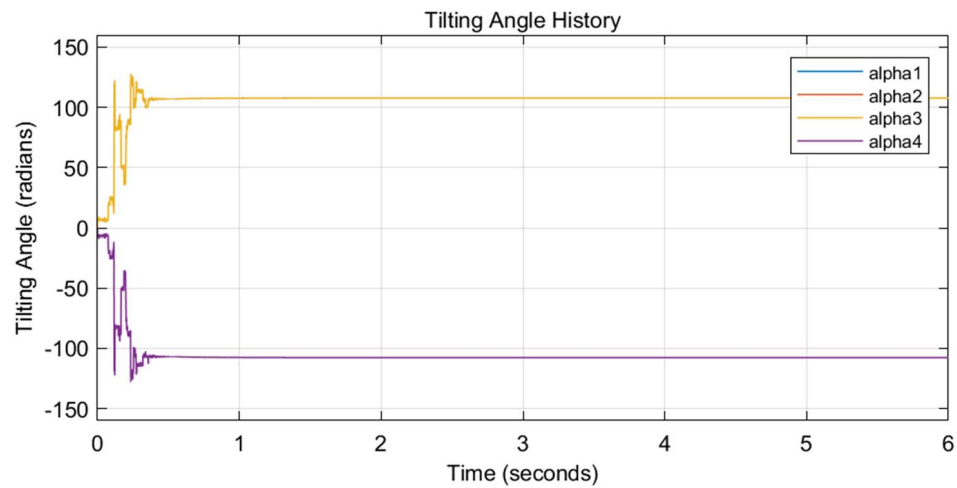

**Figure 27.** Tilting angle history of over-actuated control.

Though the response was faster than the control method proposed in this research, the tilting angles changed over-rapidly and intensively, which is commonly believed to be unrealistic.

## 7. Conclusions and Discussions

The necessary conditions to receive an invertible decoupling matrix for the attitude-altitude-based feedback linearization method were found and visualized in the tiltrotor.

For the gait restricted by $\alpha_1 = \alpha_3, \alpha_1 = \frac{1}{2} \cdot \alpha_3, \alpha_1 = -\alpha_3, \alpha_1 = -\frac{1}{2} \cdot \alpha_3$ within the specific region of interest given in (22), the feedback linearization had great success in stabilization if and only if $(\alpha_2,\alpha_4)$ lay inside the relevant region defined in Figures 20–23.

The angular velocities in each experiment, leading to a stable result, avoided touching the non-negative constraints.

The over-actuated controller in completing the same task maneuvered the tilting angle over-rapidly and intensively. In return, the tiltrotor received a faster response and less overshoot.

For the defined $(\alpha_1,\alpha_3)$ of interest, the control result is very likely to be unstable if $(\alpha_2,\alpha_4)$ lies outside the relevant colored area in Figures 20–23.

These conclusions can be effective guidance for designing the gait in a safe (invertible) region. For example, a further feedback-linearization-based gait may let $(\alpha_2,\alpha_4)$ lie within the triangular region governed by $(-1.3,1.3)$, $(1.3,-1.3)$, and $(1,1)$, since this region is mostly applicable to receiving a stable result.

The $\alpha_2 - \alpha_4$ planes (Figures 3, 5, 7, and 9) in Section 5.2 were divided into several regions by the invertibility violating curves. We argued that the applicable gait should lay in the same region. Further, we asserted that crossing these curves is inevitable when switching from the gait in one region to the gait in another region. However, this was not always true.

It can be proved that the decoupling matrix for the gait $(\alpha_1,\alpha_2,\alpha_3,\alpha_4) = (0,0,0,0)$ is always invertible, introducing no invertibility-violating curves. Thus, one may avoid crossing the invertibility-violating curves while switching the gait by adjusting $(\alpha_1,\alpha_3)$ at the same time, e.g., the tiltrotor may return to the safe gait $(\alpha_1,\alpha_2,\alpha_3,\alpha_4) = (0,0,0,0)$ as the middle step before switching to a desired gait. Further discussions are beyond the scope of this research.

Our further step is to develop a periodical gait for the tiltrotor to track a more complicated reference.